# A Simple and Practical Approach to Improve Misspellings in OCR Text

**Junxia Lin**
Georgetown University Medical Center, Georgetown University
Washington, D.C., United States
jl2687@georgetown.edu

**Johannes Ledolter**
Tippie College of Business, University of Iowa
Iowa City, IA, United States
johannes-ledolter@uiowa.edu

**Abstract**

The focus of our paper is the identification and correction of non-word errors in OCR text. Such errors may be the result of incorrect insertion, deletion, or substitution of a character, or the transposition of two adjacent characters within a single word. Or, it can be the result of word boundary problems that lead to run-on errors and incorrect-split errors.

The traditional N-gram correction methods can handle single-word errors effectively. However, they show limitations when dealing with split and merge errors. In this paper, we develop an unsupervised method that can handle both errors. The method we develop leads to a sizable improvement in the correction rates.

This tutorial paper addresses very difficult word correction problems – namely incorrect run-on and split errors – and illustrates what needs to be considered when addressing such problems. We outline a possible approach and assess its success on a limited study.

## 1. Introduction: Problem Formulation

Historic texts from the 19[th] century are typically printed in hand-carved and hand-set type. This artisanal method of type means that no two words of the same document are exactly identical in printed image. This presents a challenge for Optical Character Recognition (OCR), a common method for transferring printed text into machine editable format. While OCR has greatly improved over the last decade, it is still not completely reliable. In fact, OCR usually creates "dirty" text, with many misspellings. The word error rates of OCR output vary greatly; anywhere from 9% to 27%, depending on the type of source text (Nguyen et al (2019)). Post-processing of OCR output is essential for improving the quality of the text before it can be used for computerized text analysis.

Errors include both non-word errors and real-word errors; see Kukich (1992a). A non-word error in the OCR output is a string of letters that cannot be found in a comparison dictionary of genuine correct words. An example is the incorrect string "wcnd" in place of the intended string "word" which is an entry in the dictionary of correct words. A real-word error is the incorrect use of a correct word from the dictionary of words. As example, both the strings "word" and "work" are valid words in the dictionary; while "word" may have been the intended term, the term "work" was the term read by OCR. Real-word errors are difficult to spot. The only way to resolve real-word errors is to parse the sentence and read the text in its context.

The focus of our paper is on the identification and correction of non-word errors. Such errors may be the result of incorrect insertion, deletion, or substitution of a character, or the transposition of two adjacent characters within a single word (Damerau (1964)). Because this issue results in an error of a single word, we also refer to it as an individual word error. In addition, Kukich (1992a) points out that non-word errors can be the result of merging correct words together (by deleting the space between two or more correct words); such error is called a run-on error. Or, it can be the result of splitting a correct word into two or more parts by incorrectly inserting spaces between words; such error is called an incorrect-split error. Spaces (or blanks) are used to define the boundary of words. Hence these two errors, the run-on error and the incorrect-split error, are due to word boundary problems; the absence of empty spaces or the presence of extra spaces cause a wrong recognition of words.

Incorrect-split and run-on errors are a frequent problem. Kukich (1992b) concludes that among human-generated spelling errors, 13% of non-word error are run-on errors, while 2% are incorrect-split errors. Nguyen et al (2019) conclude that 17 percent of OCR errors are caused by word boundary problems. From their study of four large OCR datasets that represent different text sources and contain a wide range of OCR errors, they find that an incorrect-split error occurs about 2.4 times as often as a run-on error. Their result differs from that of Kukich (1992b) who claims that for human-generated misspellings run-on errors are 6.5 time more frequent than incorrect-split errors. The study by Nguyen et al (2019) also finds that incorrect-split and run-on errors are unlikely to occur together; across their four data sets only about 6.8% of errors are caused by a combination of incorrect-split and run-on errors. However, one must keep in mind that all error rates quoted in the literature depend on the quality and the characteristics of the original source text. Our OCR text was obtained from hand-carved and hand-set type from the 1860s, and we expect even higher error rates than those found by Nguyen et al.

Due to the difficulty of handling the word boundary problem, the incorrect split and run-on errors remain one of the most difficult part in the detection and correction of OCR errors. Therefore, more effective approaches are needed to handle the errors caused by the word boundary problem. The overall correction rate of OCR errors can be improved substantially once these word boundary problems can be fixed. In this paper, we propose an unsupervised OCR correction approach that corrects not only the misspellings of single words, but also the errors from words that are incorrectly merged and split.

## 2. Prior Literature

Several different post-processing OCR techniques have been developed to correct errors in OCR output text. Most techniques focus on correcting individual word errors. Many of these techniques work quite well achieving correction rates of 70% or higher. However, only few techniques exist for incorrect split and run-on errors, and their correction rates are not as clear-cut. The exponentially increasing number of possible word combinations complicates the solution of this very difficult problem (Kukich (1992a)).

Elmi and Evens (1998) propose a 3-step correction method for word boundary errors. The first step splits the input error into two strings. All possible splits are considered. When a split yields two valid words, then these two words replace the input error and the correction process ends. When the split does not produce two valid words, the procedure proceeds to the second step which concatenates the input error with the following word. If the resulting string is a valid word, the string replaces the input error and its subsequent word. The process goes to the third step if this step does not yield a valid word. The third step concatenates the previous word with the input error; if the resulting string is a valid word, the resulting string replaces the previous word and the input error.

Verberne (2002) discusses why incorrect-split and run-on errors make the detection and correction of errors so difficult. A run-on error can easily be detected (because the word is not found in lexicon), but it is difficult to correct. For correction, white spaces could be added at any position to split the error string into pieces. However, this generates a large number of possible combinations and thus reduces the speed of correction. The incorrect-split error is difficult, for both detection and correction. Pieces of the incorrect split often contain one or more correct words that can be found in the lexicon. This increases the difficulty to detect this type of error because individual split pieces are already identified as correct words. If neighboring words are not taken into account, the incorrect split-error cannot be corrected properly. But taking the neighboring words into consideration again reduces the speed of detection and correction.

Kolak et al (2003) propose a channel-based approach that incorporates a probabilistic segmentation model to handle word merge and word split issues. Several sets of boundary positions are applied to each sequence of multiple words. Based on the selected positions, each word sequence is segmented into multiple components, and the components are checked for non-word errors. While this method reduces word error rates considerably, its segmentation process requires expensive computation as there are many possible sets of boundary positions.

Evershed and Fitch (2014) propose a method that generates correction candidates by considering the left and right neighboring words of the error.

Afli et al (2015) utilize a statistical machine translation system (SMT) to translate the OCR output text into its corrected form. Their system uses a statistical model that relates features in the OCR output text to a manually corrected version. This method has the potential to handle the incorrect-split and run-on errors in the OCR text, because the OCR text and its correct version are trained for aligned sequences, instead of on single words separated by spaces. Statistical features between the true text and its incorrect-split or run-on errors are learned along with other errors, and these features enable the model to correct incorrect-split and run-on errors. However, this method requires an OCR version without errors to train the data. Furthermore, for the training step the OCR output and its correct version need to be carefully aligned.

Mei et al (2017) integrate a word segmentation step into their correction method. Prior to the error detection step on individual words, they employ Google N-gram tokenization methods to split tokens and then use heuristics rules to merge the incorrect split fragments into a single unit.

The OCR post-correction system discussed by Schulz and Kuhn (2017) considers a combination of several correction techniques – including statistical machine translation, the spell checker hunspell[1], and word compound-splitter tools – to suggest candidate corrections for errors in an input OCR text. A ranking algorithm is then applied to select the best-fitting correction candidate for the word error. The ranking algorithm makes use of a language model and a phrase table that is generated from all input OCRed text and all candidate corrections when assigning weights to the set of correction techniques.

Dong and Smith (2018) utilize duplicate OCR texts, if available, to build an error correction model. Their proposed correction method uses multiple variants of OCR text obtained from the same original text to train the correction model and generate a better version with fewer errors. Several combination strategies are developed, and optimal weights for their combinations are developed. The authors claim that their novel multi-input approach performs as well as other available supervised correction methods.

Soni et al (2019) propose an unsupervised OCR post-correction approach for out-of-vocabulary errors that contain a missing whitespace based on Google n-gram counts. This approach corrects the error by applying two methods. One is context-free, while the other considers the context of the error. String segmentation tools are used in both methods to help split strings.

## 3. The Datasets and Set-up

The Debates of the 39th U.S. Congress, as recorded in the Congressional Globe, were the inspiration for our paper. In their recent book, Ledolter and VanderVelde (2021) analyzed the digitized proceedings of the 39th U.S. Congress that met from 1865 to 1867. This is an official account of exactly what was said and done on the floor of Congress, published for the reading of the American people and for posterity. The digitization started from the printed record of the speeches. Each page was subjected to OCR and converted into an electronic record and stored in Microsoft Word format. Pages were combined into a giant master file, all page breaks, headers, footers, and page numbers were stripped off, and separation symbols were entered to indicate the start of each speech. The correction of the numerous OCR errors was tedious, lengthy and frequently incomplete. The challenge with "cleaning" some of these errors has led us to this tutorial paper.

In this paper, we propose an automatic correction method for non-word errors in OCR text that also takes into account incorrect-split and run-on errors. A brief outline of what we do is given below.

---

[1] https://github.com/hunspell/hunspell

For error detection and correction we use the Corpus of Historical American English (COHA), a large comparison corpus with its dictionary of true words. The full COHA contains texts of fiction and non-fiction, magazines and newspapers from the 1810s to 2000s, and includes more than 400 million words (Davies (2010)). We work with the freely-available sample COHA that represents a 1/100th random selection of the full COHA corpus. We assume that relative frequencies of N-grams in the sample corpus approximate relative frequency in the full corpus. We select the large COHA corpus as reference because the OCR text in need of word correction represents historical material from the 19th century. Taking a reference corpus that covers this time period reduces the likelihood that a 19th century word that has disappeared in modern times is not recognized as a correct word.

The OCR text is tokenized into words, treating white spaces as word boundaries. The same steps are applied to the sample COHA reference corpus. Its dictionary consists of all words that occur at least 5 times. The reason for excluding very low frequency words is that such words are usually not real words and not helpful for correction. Furthermore, all words with two or fewer characters are excluded from the dictionary. The reason for this is that we do not correct words of lengths 1 and 2 in the OCR text as short words are usually omitted from the document-term frequency matrix. The document-term frequency matrix is the starting point for most textual analysis; words of length 2 or shorter do not matter, whether they are correct or misspelled.

The dictionary is used to detect non-word errors. Any word in the OCR text of length 3 or higher that cannot be found in the reference dictionary of real words is flagged as a non-word error. These are the words that need to be corrected.

## 4. The Proposed Method

### 4.1  Ignoring word boundary problems

Initially, we ignore word boundary problems. For each non-word error, we calculate its Damerau–Levenshtein distance to all words in the comparison dictionary. The DL distance (Damerau (1964), Levenshtein (1966)) measures the edit distance between two character-sequences. The DL distance between two words is the minimum number of operations required to change one word into the other. Permissible operations consist of insertions, deletions or substitutions of a single character, or transposition of two adjacent characters. Efficient iterative algorithms are available to calculate this distance. Since the number of words in the reference dictionary is huge, any procedure that calculates pair-wise distances must be computationally efficient and fast.

To explain our procedure more fully, let us consider a certain non-word error. This string is compared with each of the many words in the reference dictionary. A list of correction candidates is generated from the DL distances between the non-word error and each word in the dictionary. A threshold on the DL distance is applied; only COHA words with DL distances at or below a certain threshold become candidates for correction. No corrections are made if the closest word match with the word-error has a DL distance larger than this limit. For errors with word lengths between 3 and 5, the threshold is set at 1. If there are no candidates with DL=1, the error is treated as uncorrectable and will not be processed to a correction. For errors with word lengths between 6 and 9, the threshold is set at 2. For errors with word length 10 or greater, the DL threshold is set at 3. The error is treated as uncorrectable if no candidate words with DL <= 3 can be found.

If there are correction candidates, we proceed to the correction step. Candidates are processed for correction according to their DL distance. The frequencies of the terms in the candidate set from the reference dictionary determine the correction: the candidate word with the largest frequency in the reference dictionary becomes the correction. If we have candidates with DL=1, then the best of these candidates becomes the correction. If there are no words with DL=1, then we consider the candidates with DL=2 and make the correction, and so on until we reach the DL threshold. This is the very simplest method which only looks at dictionary words (or 1-grams). This is our Method 1.

We also investigate whether an extension to N-grams is useful. There we look at the adjacent words (one word before, or two words before, or the two words adjacent) to the non-word error in the OCR text, and form bigrams (or trigrams, if two additional words are considered) with all candidate replacements from the reference dictionary. The most frequent bigram (or trigram) determines the correction. The replacement candidate in the most frequent bigram (trigram) becomes the correction of the non-word error. This general method is referred to as the N-gram language model correction technique. These are our Method 2 (with the word before), Method 3 (with two words before), and Method 4 (with the two adjacent words), respectively.

Smoothing is not incorporated in our N-gram correction methods. Smoothing, such as Laplace Smoothing, is commonly used in situations when counts of n-grams are too small to calculate the probabilities of all candidate n-grams. Our methods do not require the calculation of all these probabilities. Our purpose is simply to determine frequencies, rank the list of candidate n-grams, and pick the top ones.

## 4.2 Addressing word boundary problems

A novel 4-step approach is used to correct incorrect-split and run-on errors.

**Step 1** includes three sub-steps.

The **first sub-step** is about error detection and is the same as described earlier. We detect non-word errors by comparing each word in the OCR text with all words in the reference dictionary. An error is found when there is no match in the reference dictionary.

The **second sub-step** embeds errors into a string without blanks (spaces). The generation of this zero-blank string depends on the error pattern of consecutive words. We distinguish two error patterns. The first pattern is one when neighboring words of an error are both correct words. In this situation, the zero-blank string is generated from the error itself. For example, take the OCR text "in refereneeto these" where "refereneeto" is a detected error and both its neighboring words, "in" and "these", are both correct. The zero-blank string is therefore "refereneeto". The second error pattern is one when two or more errors occur consecutively. In this case, the zero-blank string is generated by concatenating the first two errors together joining them with a special segmentation symbol such as the underscore "_". For example, consider the OCR text "in refe rene eto these" in which "refe", "rene", and "eto" are consecutive errors. Then, a zero-blank string is generated by concatenating the first two errors, "refe" and "rene". The errors are joined with the segmentation symbol, resulting in "refe_rene". The reason for using a segmentation symbol is described below. The reason why we concatenate only the first two errors (and not three or more) has to do with computing efficiency.

The **third sub-step** creates string-splits of the resulting zero-blank string. The string is split into a set of correct words and word-error combination through a novel specially-designed depth-first search (DFS) algorithm. The segmentation symbol "_" is added to the comparison dictionary to help us split the zero-blank string at its original position. Our string-split algorithm relies on the words in the comparison dictionary when cutting off substrings; by adding the segmentation symbol to the dictionary, the string-split algorithm is able to cut off "_" and separate the preceding and successive substring around it. This helps us get back to original text of the zero-blank string. The reason why we want to get back to the original text behind the zero-blank string is that the two consecutive errors in the original OCR text could be two components of a split word, such as "comp onents" split from "components", but could also be two single word errors, such as the two errors "smgle wond" for "single word". By allowing the string-split algorithm to split the zero-blank string at the segmentation symbol, both situations can be handled.

Our DFS is able to help us identify for a zero-blank string all possible combinations of correct words and non-word errors. The DFS string-split algorithm starts its search at the first character of the string and adds the following characters to form a substring; once a substring can be matched to a real word in the comparable lexicon, the substring is cut off from the rest of the string, and the remaining string is searched the same way. After this first round, the search starts at the second character of the string and repeats the same process. And so on. The search process ends when the starting point of a round reaches the last character of the string.

The decision in the second sub-step to concatenate only the first two errors was made because most incorrect-split words are split into just two components; only few are split into more than two. Another reason is computation, which is important if large texts need to be corrected. Concatenating only two errors yields a zero-blank string of manageable length (e.g., less than 15 characters) and the subsequent split into its combinations will be fast. If we have a long string (e.g., 20 characters), the time of splitting increases exponentially.

Consider the zero-blank string "cont_estedelection" as an illustrative example. Assume that the dictionary consists of the following 6 terms: "contested", "election", "test", "tested", "contest", and the added segmentation symbol "_". The third sub-step applies the DFS split and generates the following set of combinations:

[1] "cont _ ested election"
[2] "cont _ estedelection"
[3] "cont_ested election"
[4] "cont_estedelection"

There are 4 combinations in this example. The first "cont _ ested election" is the combination of "cont", "_", "ested", and "election". The terms "_" and "election" are correct words because they are in the dictionary. The other terms are non-word errors because they are not in the dictionary. The second "cont _ estedelection" is the combination of "cont", "_", and "estedelection". Only the term "_" is a correct word because the segmentation symbol is part of the dictionary. The third "cont_ested election" is the combination of "cont_ested" and "election". The term "election" is a correct word found in the dictionary. The last "cont_estedelection" is the combination of the zero-blank string itself. When the string-split gets to the end of the search, no split is performed, and the whole zero-blank string is returned.

**Step 2** involves the correction of all single word-errors in each of the word combinations that have been generated by the previous step. We use the N-gram correction method (Method 4 with two adjacent words) that we described previously. All remaining segmentation symbols are either removed or replaced by a space. The segmentation symbol within corrected single word-errors disappear during the N-gram correction process. However, it will be left in uncorrectable errors, such as "cont_estedelection", or when it appears as a separate symbol, such as "cont _ ested election". A segmentation symbol within an uncorrectable error such as "cont_estedelection" is replaced by a space. A separate segmentation symbol, such as the one in "cont _ ested election", is removed because it is unnecessary.

For the illustrative example, Step 2 leads to the following corrected combinations; words in boldface are the corrected words:

[1] "cont **tested** election"
[2] "cont estedelection"
[3] "**contested** election"
[4] "cont estedelection"

**Step 3** evaluates the correction of each word combination. We consider three measures. First, the number of remaining errors, which is defined as the number of uncorrectable errors in each combination. In our illustration, there is one remaining error in the first combination ("cont"); two remaining errors in the second combination ("cont" and "estedelection"); no errors in the third combination; and two errors in the last combination ("cont" and "estedelection"). Second, the DL distance between the original OCR string and the combination string. In our example, the original OCR string is "cont estedelection". It is compared to each of the four corrected combinations from Step 2, resulting in four DL distances. The third measure is based on a fitted topic model. The Latent Dirichlet allocation (LDA) algorithm is used to fit 5 topics to the words of original uncorrected OCR text. Each topic assigns to words probability scores that indicate how relevant words are to the topic. For each combination we sum its matched word's probability scores that are assigned by each of the five topics. The largest sum from these five topics expresses the quality of the correction of that combination.

**Step 4** finalizes the correction as the correction with the fewest number of errors, with ties broken by the lowest DL distance first, and the largest probability scores second.

## 5. Results

We select 50 sample speeches from our OCR text corpus to examine the correction accuracy of the methods. The sample speeches contain about 2,300 words. The total number of detected non-word errors with word length 3 or higher is 235. We use two metrics to examine the methods' correction accuracy:

1. The uncorrectable error rate, defined as the number of uncorrectable errors divided by the total number of detected non-word errors with word length 3 or higher.

2. The appropriate ("right") correction rate, defined by the number of appropriate corrections divided by the total number of detected non-word errors with word length 3 or higher. The appropriateness of the correction (whether a correction is in fact right or wrong) is assessed subjectively by reading each sentence in its context.

Results are summarized in Figure 1. For methods that ignore word boundary problems (Methods 1 through 4), 82 out of 235 non-word errors are uncorrectable; the uncorrectable error rate is 34.9%. The remaining 153 errors are corrected and replaced by real words. We then assess these 153 corrections (checking whether there are actually right or wrong) by reading each sentence in its context. For Methods 1 through 4, the numbers of right corrections are 78, 81, 81, and 82, respectively, which imply 33.2%, 34.5%, 34.5%, and 34.9% right correction rates.

Method 5 addresses word boundary problems; 39 out of 235 non-word errors are uncorrectable, for an uncorrectable error rate of 16.6%. The remaining 196 errors are corrected and replaced by real words. Inspecting the 196 corrections we find that number of appropriate corrections is 133, implying a 56.6% right correction rate. The 133 right corrections also include 8 partially right corrections. Partially right corrections are corrections where an original error in the OCR text such as "notedbycongressitproposed" is corrected to "note day congress proposed." The correction is partially right as in this case "congress" and "proposed" are recognized correctly.

In conclusion, the correction accuracy of Method 5 is 56.6%, a significant improvement when compared with methods that ignore boundary problems.

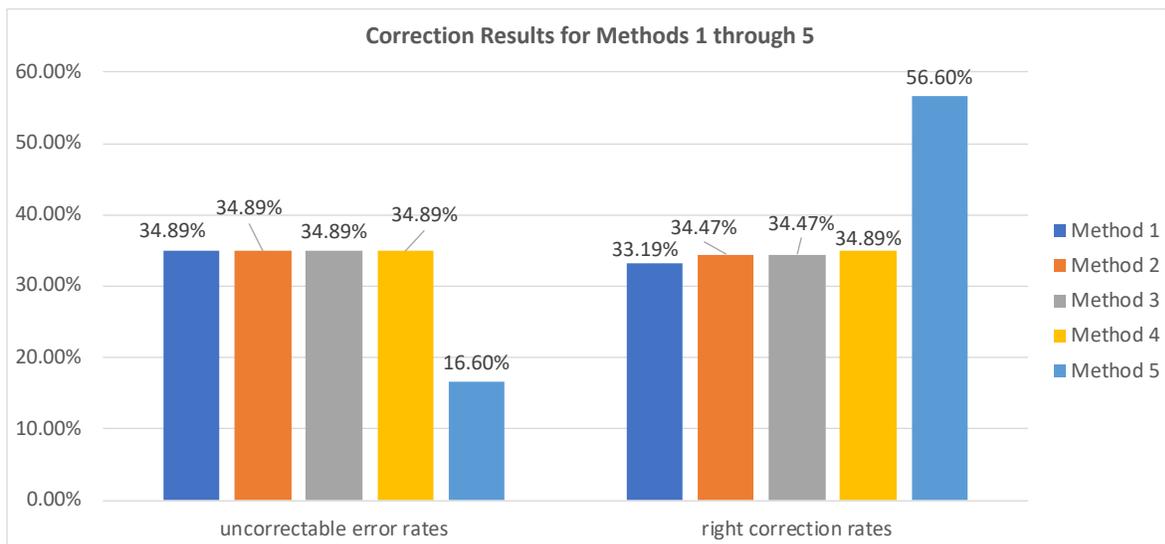

**Figure 1:** Uncorrectable error rates and right correction rates for Methods 1 through 5.

## 6. Discussion

The uncorrectable error rates do not differ much between the first four methods, and is about 35 percent. This is because all four methods generate the candidates for correction the same way and they do not consider word boundary problems. The uncorrectable error rate for Method 5 decreases sharply to 16.6 percent.

The right correction rates increase slightly when the correction considers neighboring words around the error, from about 33 percent for the simplest Method 1 to about 35 percent for Method 4 that considers both words around the error. The right correction rate increases to 56.6 percent for Method 5 that incorporates incorrect-split and run-on errors. This is a large improvement. N-gram correction models can handle single word error effectively. However, they show limitations when dealing with split and merge errors. Our Method 5 leads to a sizeable improvement.

The difficulty to handle exponentially increasing numbers of possible word combinations (mentioned by Kukich (1992a)) has been one of the biggest challenges for the correction of boundary word problems. Our Method 5 presents a practical solution to this challenge. It utilizes an effective string-split technique to correct a string that is compromised by boundary problems. Our procedure includes several steps, and each of these steps could be optimized even further. Method 5 is a useful approach to word error correction if the OCR output contains a considerable portion of word split and merge errors.

Computer running time is an important issue for many correction methods, especially if they include complicated algorithms. Our programs are written in the computer language R. Methods 1 through 4 takes less than a minute on our document with 2,300 words and about 230 errors to correct. Method 5 takes about 20 minutes.